\documentclass[letterpaper, 10 pt, conference]{ieeeconf}  

\IEEEoverridecommandlockouts                              
\usepackage[utf8]{inputenc}
\usepackage[T1]{fontenc}
\usepackage{amsmath}
\usepackage{float}

\usepackage{graphicx} 

\title{\LARGE \bf An Energy-Efficient Reconfigurable Autoencoder Implementation on FPGA}

\author{Murat Isik\(^\dagger\), Matthew Oldland\(^\dagger\), Lifeng Zhou
\thanks{\(^\dagger\)The authors contributed equally to the work.}
\thanks{All authors are with Electrical and Computer Engineering, Drexel University.}
\thanks {Github Link: https://github.com/Muratcanisik4/CNN-Autoencoder.git}
}

\begin{document}

\maketitle
\thispagestyle{empty}
\pagestyle{empty}

\begin{abstract}

Autoencoders are unsupervised neural networks that are used to process and compress input data and then reconstruct the data back to the original data size. This allows autoencoders to be used for different processing applications such as data compression, image classification, image noise reduction, and image coloring. Hardware-wise, re-configurable architectures like Field Programmable Gate Arrays (FPGAs) have been used for accelerating computations from several domains because of their unique combination of flexibility, performance, and power efficiency.  In this paper, we look at the different autoencoders available and use the convolutional autoencoder in both FPGA and GPU-based implementations to process noisy static MNIST images. We compare the different results achieved with the FPGA and GPU-based implementations and then discuss the pros and cons of each implementation. The evaluation of the proposed design achieved 80\% accuracy and our experimental results show that the proposed accelerator achieves a throughput of 21.12 Giga-Operations Per Second (GOP/s) with a 5.93 W on-chip power consumption at 100 MHz. The comparison results with off-the-shelf devices and recent state-of-the-art implementations illustrate that the proposed accelerator has obvious advantages in terms of energy efficiency and design flexibility. We also discuss future work that can be done with the use of our proposed accelerator. 

\end{abstract}

\section{INTRODUCTION}

Autoencoders are an unsupervised neural network that aims to learn how to compress input data and then reconstruct the data to match the input data with the least amount of loss \cite{c1}. This is done using encoder and decoder methods, where the encoder compresses the data and the decoder reconstructs the data \cite{c1},\cite{c2},\cite{c3},\cite{c7}. There are many types of autoencoders that can be used in machine learning; some examples are simple, convolutional, removing noise, and sparse autoencoders. These different autoencoders can then be used for different processing needs, like data compression before storage, image classification, and removing noise from images. 

This paper shows the use of convolutional autoencoders and how effective they are at removing noise from images. Convolutional autoencoders use convolutional neural networks for the layers of encoding and decoding to help compress and reconstruct the image respectively. Convolution neural networks use different filters to pull out features from the data. These filters are designed to pull out specific features and are dependent on what the feature needed is. The filters are designed to scan through the data and then create a map of how well the data matched the filter by creating scores, where a higher score is a better match \cite{c4}. The filter strides across the image based on how the layer is set up by the user. These maps are then used in the next layer, which is either a pooling layer or an up-sample layer. The pooling layer looks at a window of the data and will combines the window into one sample. The pool can use either a max of the data, the sum of the data, or the average of the data to create the new sample \cite{c4}. The pooling layer is used in the encoding process to help compress the data and store the important features. While the up-sampling process is used in the decoding process to help with reconstruction. There are many different ways to implement an autoencoder using different hardware. This paper compares two methods, one is an implementation of an FPGA using VHDL and the other is a GPU-based implementation using python. 

While developing the rest of the framework, machine vision framework developers and respectability might become engrossed in deciding which of these steps to use. The organize option is commonly chosen when prototyping a system for the first time. A prototype application must be tested to determine how many sections it must prepare every second or how many outlines it must handle per second of the live video. The power consumption of embedded systems is an important consideration for most applications, and researchers strive to develop energy-efficient design methodologies [14-15].

The applications require for handling enormous data sets of high-resolution image preparing applications requests faster, configurable, long throughput frameworks with superior vitality productivity [16-17]. FPGAs (Field-Programmable Gate Arrays) can play an important role since they provide configurability, adaptability, and parallelism to coordinate the necessary throughput rates of the application under consideration \cite{c13}. Real-world applications can be implemented on FPGA devices thanks to their execution capabilities. The development of FPGAs has taken hardware flexibility, in general, one step further. In the end, The toolchain for developing applications on these devices has also advanced significantly, allowing these devices to be used by a larger building community \cite{c18}. FPGAs are typically used in applications that demand concurrency, high transfer speeds, and re-programmability.

FPGAs have achieved rapid acceptance and growth over the past decade because they can be used in a very wide range of applications. The algorithms that extract features are time-consuming, which is a huge drawback when developing real-time applications. One solution to this problem is the use of dedicated hardware for the algorithms, such as the FPGA, which can provide dedicated functional blocks that perform complex image processing operations in parallel. FPGAs have achieved rapid acceptance and growth over the past decade because they can be used in a very wide range of applications \cite{c19}.

\vspace{.5cm}We designed the convolutional autoencoder in both FPGA and GPU-based implementations to process noisy static MNIST images. Results show that the proposed design is both area- and power-efficient.

\section{AUTOENCODER TYPES}

Autoencoders are a type of unsupervised neural network and are used for various processing applications. They are trained to encode and compress input data then reconstruct and decode the data back to try to match the input data. There are three main parts to autoencoders \cite{c3}:

\begin{itemize}
    \item Encode and Compress,
    \item Bottleneck,
    \item Decode and reconstruct.
\end{itemize}

The encode and compress part of the autoencoder is composed of layers of neural networks that work to compress the data down to a set size. This encoded and compressed data is considered the bottleneck part since it is the smallest part of the system before getting reconstructed. The data could then be stored if needed, which allows for less required storage capacity. The compressed data is then used in the next part of the system, the decoding, and reconstruction part. This part works to decode the encoded data and reconstruct the data back to match the input data that was passed in. It does this by using neural networks that are designed the same as in the encode and compress part, however instead of compressing the data it is up-sampled instead. The number of layers used in both the compression and reconstruction parts is set by the user based on the required implementation needed. Figure 1, below, shows an example of an input image being compressed and reconstructed to an output image that matches closely \cite{c2}. 

\begin{figure}
      \centering
      \includegraphics[scale=.4]{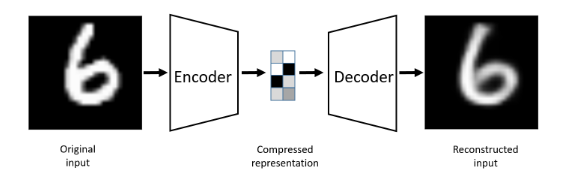}
      \caption{Example Autoencoder showing input image compression and reconstruction \cite{c1}}
      \label{figurelabel1}
\end{figure}

There are many different types of autoencoders that are used within machine learning. These autoencoders have the same flow and parts between them but the layers used within the encoding and decoding change. The sections below cover some of the different types of autoencoders that can be used. 

\subsection{Simple Autoencoder}

Simple autoencoders are the basic implementation of an autoencoder and are a simple feed-forward network that can have multiple layers within the encode and decode parts. One implementation had 3 layers and the layers were fully connected to each other \cite{c1}. This implementation also had the input and out of the autoencoder as the same size which forces the autoencoder to learn the compressed data with no information loss \cite{c1}.

\subsection{Convolutional Autoencoder}

Convolutional autoencoders are similar to simple autoencoders but change the layers to convolution layers using a convolution neural network. The convolution neural network layers are set up with several parts \cite{c4}:
  
\begin{itemize}
    \item Convolution,
    \item Rectified Linear Unit,
    \item Pooling (Encode) or Up-Sample (Decode).
\end{itemize}

The convolution is performed by forming features out of the data and creating feature maps using convolution window filters. These filters are used to maintain the relationship between the pixels and the input image \cite{c4}. The resulting features create high and low scores that are representative of how well the pixels are in relation to the filter; the better the match the higher the score. The convolution has two parameters to it, the padding and the stride length \cite{c4}. The padding is so that feature has the same amount to work on no matter the input data size. The stride parameter is set to determine how many pixels the filter shifts over the input data when creating the feature. 

The Rectified Linear Unit (ReLU) part is used to make sure that there are no negative numbers going between layers. This is done by replacing any negative number found with zeros to help the math work out. The pooling part is the compression of the image. A window size is selected to combine parts of the filtered features into one pixel of the new image to be passed to the next layer. In the pooling window either the max value, average value, or summed value within the window is passed into the next pixel of the new image. The decoding and reconstruction layers will use an up-sample part instead of the pooling part to increase the size of the image as it gets reconstructed back to match the input size.

\subsection{Denoising Autoencoder}

Denoising autoencoders \cite{c2} use Gaussian noise on the input data to disrupt the data then the autoencoder is trained to reconstruct the data with the noise removed. The goal of the denoising autoencoder is to be used for error correction of data that has been corrupted. The effectiveness of the autoencoder is measured by the reconstruction loss between the output data and the input data. 

\subsection{Sparse Autoencoder}

Sparse Autoencoders use sparsity in the layers to help with keeping the reconstruction error low and have a meaningful bottleneck image in the middle. In \cite{c2}, the sparsity was enforced by applying L$_1$ regularization in the layer. This changes the autoencoder optimization objective to:
\begin{equation}
    \text{argmin}_{A,B} E[\Delta(x,B\circ A(x)] + \lambda\sum_{i}|a_{i}|,
\end{equation}
where a$_i$ is the activation at the $i$th hidden layer and $i$ iterates over all hidden activations \cite{c2}. Another method covered in \cite{c2}, is the KL-divergence method. This method measures the distance between two probabilities by changing the activation a$_i$ to a Bernoulli variable with a probability of $p$ \cite{c2}. The method tweaks the probability $p$ by taking the measured probability for a batch, finding the difference, and then applying that as the regularization factor. 

\section{Method}

We consider an image has been corrupted by noise and requires the noise to be removed for processing later on. A convolutional autoencoder was designed to perform image noise reduction on the corrupted input images. The convolutional autoencoder is designed using three convolution layers and three pooling layers, interleaved, for encoding and compression. The decoding and reconstruction used four convolution layers and three up-sampling layers that were interleaved. The pooling layers were constructed to use the maximum pooling operation when operating over the windows of the images. 

\begin{table}
\centering
    \begin{tabular}{| c | c | c | c |} 
     \hline
     Layer & Type & Window Size & Stride \\ [0.5ex] 
     \hline\hline
     1 & Convolution & 4 & 1 \\ 
     \hline
     2 & Max Pool & 4 & 1 \\
    \hline
     3 & Convolution & 4 & 1 \\ 
     \hline
     4 & Max Pool & 4 & 1 \\
     \hline
     5 & Convolution & 4 & 1 \\ 
     \hline
     6 & Max Pool & 4 & 1 \\
     \hline
     7 & Convolution & 4 & 1 \\ 
     \hline
     8 & Up-Sample & 4 & 1 \\
     \hline
     9 & Convolution & 4 & 1 \\ 
     \hline
     10 & Up-Sample & 4 & 1 \\
     \hline
     11 & Convolution & 4 & 1 \\ 
     \hline
     12 & Up-Sample & 4 & 1 \\
     \hline
     13 & Convolution & 4 & 1 \\ [1ex] 
     \hline
    \end{tabular}
\caption{Layer Construction for Convolutional Autoencoder}
\label{table:table_1}
\end{table}

Table \ref{table:table_1} shows the layer construction of the convolutional autoencoder for encoding and decoding.

The convolution layers were designed using convolution neural networks to train the weights and perform feature extraction on the image before the pooling or up-sampling layers. A 2-D convolution was used as the filter to create the features from the pixels. The general 2-D function equation used is given by
\begin{equation}
    f = \sum_{a=-y}^{y} \sum_{b=-z}^{z} x_{i-a,j-b}*k_{a,b}.
\end{equation}

After the convolutional filters are run over the pixels a ReLU function is performed to make sure that no negative numbers are passed to the next layer. This allows for the math to always come back correct and not have errors when performing calculations. The ReLU function used is given by 
\begin{equation}
    y = \max(0,x),
\end{equation}
where x is the filtered pixel feature. The output of the ReLU function is then passed to the next layer of either pooling or up-sampling. The pooling function was a max function of window size of the filtered pixel features and is given by
\begin{equation}
    y = \max(x_1,1,...,x_{i,j}),
\end{equation}
where $i$ and $j$ are the window size chosen for the layer. 

\section{RESULTS}

\subsection{Dataset}

This dataset is to be a noisy variation of the MNIST digit set, with the noise coming from an internal sense of style. The training set receives 80\% of the whole data, whereas the validation set receives 20\% of the total data. The actual training data contains 60,000 photos in all. Since the dataset, we require for the task at hand requires noisy images, we define functions to add random gaussian noise to the dataset elements. The new dataset now consists of noisy images and can be visualized.
\begin{figure}[h]
    \graphicspath{ {D:\Stack} }
    \center \includegraphics[width=6cm, height=4cm]{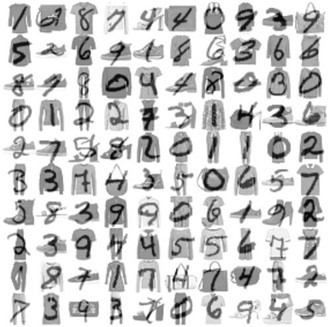}
    \caption{Project noisy dataset}
    \label{Figure 2}
\end{figure}

\vspace{.5cm}Our noisy dataset is shown Figure \ref{Figure 2}. Training data contains 60k photos in all. We use the random gaussian noise function to make the MNIST dataset into a noisy dataset.

\vspace{.5cm}Gaussian noise is a type of noise that is in the form of Gaussian distribution, such as the white noise commonly encountered. It is random-valued and in impulses.

\subsection{Training Details}

\begin{figure}[h]
    \graphicspath{ {D:\Stack} }
    \center \includegraphics[width=6cm, height=4cm]{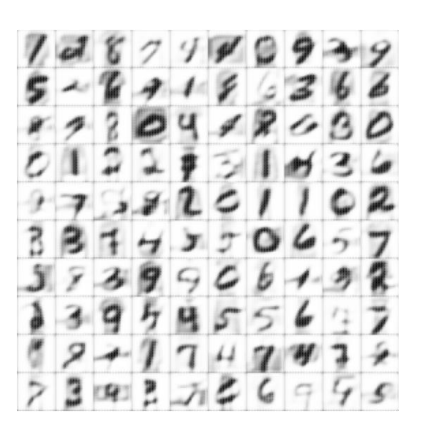}
    \caption{Denoised Image}
    \label{Figure 3}
\end{figure}

We eventually settled on the 20-epoch training time and the training loss and validation loss are displayed at the end of each epoch. Training time for each model amounted to about 100 seconds each due to the input images being of small size and hence not requiring too much processing power. Although it can occasionally reduce the visibility of already visible numerals, it does a better job of cleaning out the number and making it far more visible in more challenging circumstances like the first image. For a modest number of training parameters, the results are reasonable, with the greatest accuracy of 80.83\%. The output of our autoencoding process is shown in Figure \ref{Figure 3}.

\subsection{Experimental Setup}

We focused on two approaches to physically implementing the convolutional autoencoder; one using a GPU and the other using an FPGA hardware implementation. The GPU implementation was done using Python and Google Colaboratory. Google Colaboratory allowed for running the machine learning python executable on an NVIDIA Tesla K80. We implement the same design on the XCZU7EV FPGA chip having a 504K lookup table (LUT) and used 2021.1 software version of Vivado.

\subsection{FPGA Implementation}
The encoded images are further analyzed with a convolutional autoencoder architecture comprising of matrix operation, channel distributor, encoding module, round-robin arbitration, and output controller blocks.

\textbf{Matrix operation}: The boundary information during the convolution operation (28*28*1).

\textbf{Channel Distributor}: Matrix is sent to the channel distributor module, which contains a channel controller and first-in-first-out memories. The channel controller interleaves channel data points and then sends them to the FIFOs in sequence. The FIFOs buffer the interleaved data and then divide them into eight channels.

\textbf{Encoding and decoding module}: First, the convolutional encoding module does the convolution process, activation, and then max-pooling. Decoding is performed by up-sampling rather than max-pooling.

\textbf{Round robin arbitration}: The output data of the convolutional encoding module are orderly packaged and then sent to the output controller. Round-Robin arbitrator is giving priority to requestors. 

\textbf{Output controller}: This module is used to judge whether the data stream is processed completely.

\begin{figure}[h]
    \graphicspath{ {D:\Stack} }
    \center \includegraphics[width=8cm, height=4cm]{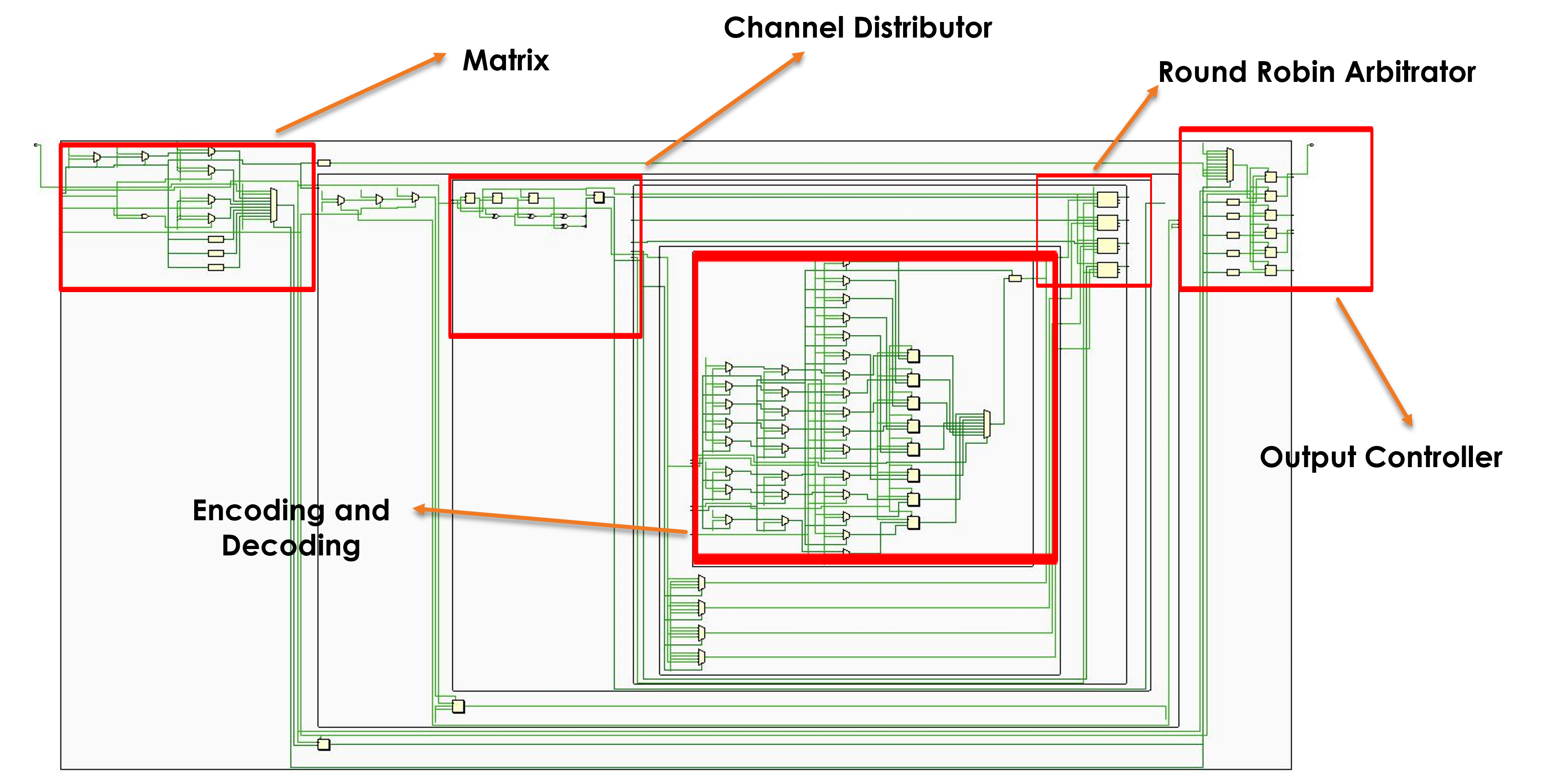}
    \caption{RTL Design of the Implementation}
\end{figure}
    
\subsection{Comparision}

\begin{figure*}
    \graphicspath{ {D:\Stack} }
    \center \includegraphics[width=0.8\textwidth]{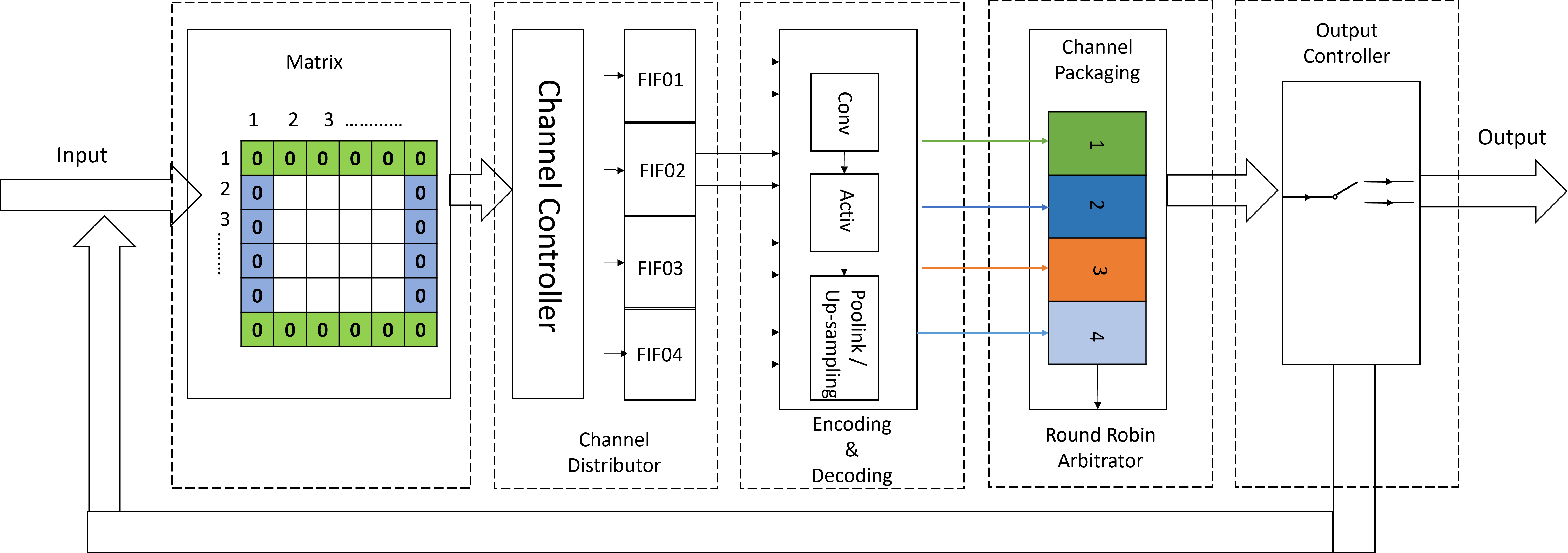}
    \caption{Block Diagram of Implementation}
\end{figure*}

\begin{table*}
    \renewcommand{\arraystretch}{1.2}
    	\setlength{\tabcolsep}{3pt}
    \centering
    \caption{Autoencoder on different heterogeneous devices}
    \label{table:table_2}
    \resizebox{0.8\linewidth}{!}{
    \begin{tabular}{llllllll}
         Platform&Technology  &Operation Frequency &Power  & Latency & Throughput &Energy Efficiency  \\\hline
         NVIDIA K80&ASIC (28nm)    &$1.48GHz$&$30W$&$1137.62
        ms$&$22000(GOP/s)$&$733.33(GOP/s/W)$ \\
        NVIDIA GTX 1080 TI&GPU(16nm)    &$1.48GHz$&$250W$&$6.15ms$&$235.77
        (GOP/s)$&$0.94(GOP/s/W)$\\
        {Chakradhar et al.~\cite{c12}}&{FPGA(28nm)}&\textbf{$200MHz$}&$15W$ &{$-$}&{$16(GOP/s)$}&$1.06(GOP/s/W)$\\
        {Gokhale et al.~\cite{c11}}&FPGA(28nm)    &$150MHz$&$8W$&$4.50ms$&$23.18
        (GOP/s)$&$2.90(GOP/s/W)$\\
        {Zhang et al.~\cite{c5}}&{FPGA(28nm)}&\textbf{$100MHz$}&$18.61W$ &{$21.61ms$}&{$61.62
        (GOP/s)$}&$3.31(GOP/s/W)$\\
        {Ours}&{FPGA(16nm)}&\textbf{$100MHz$}&$5.93W$ &{$2.91ms$}&{$21.12(GOP/s)$}&$3.56(GOP/s/W)$\\
    \end{tabular}}
\end{table*}

The implementation of Neural Networks (NNs) has been done using CPUs, GPUs, and ASICs. CPU/GPU-based NNs consume a lot of power and have a limited throughput due to limited memory bandwidth. Many researchers have developed custom ASICs for accelerating network inference workloads in order to achieve the best performance and energy efficiency. In spite of their attractiveness, ASICs cannot provide sufficient flexibility to accommodate the rapid development of Neural Networks, and FPGAs function as programmable devices that can construct unique logic, alleviating constraints on neural network implementation. As a result, one of the current research hotspots involves the development of hardware systems supporting NN inference based on FPGA to achieve high throughput and power efficiency. 

As shown in Table \ref{table:table_2}, we compare our CNN autoencoder implementation with other CNN autoencoder models presented in the literature. The evaluation of the proposed design achieved 80\% accuracy and our experimental results show that the proposed accelerator achieves a throughput of 21.12 Giga-Operations Per Second (GOP/s) with a 5.93 W on-chip power consumption at 100 MHz. The comparison results with off-the-shelf devices and recent state-of-the-art implementations illustrate that the proposed accelerator has obvious advantages in terms of energy efficiency and design flexibility.
\vspace{.5cm}
\section{Conclusion and Future Work}
In a confusion matrix, the model determined which classes it is accurately forecasting and which classes it is forecasting inaccurately. More training parameters and deeper layers in the autoencoder could improve the model's performance. However, in the context of this study, the model performs admirably in reassembling denoised photos. It is shown that the use of convolutional layers is very important in the case of denoising images using deep learning based on a distinction between encoder and decoder architectures. The model is implemented for hardware acceleration with various heterogeneous devices and resulting in an energy-efficient, reconfigurable system on the latter. Therefore, a deeper model and a more suitable latent space dimension can be used for more complex denoising involving RGB images. Future work scenarios could be removing noise from pictures received from cameras attached to autonomous robots before image processing, as the next step of this research. We are seeking to integrate our model into robots so that a response can be made by cleaning images received wirelessly from different robots if the transition is not clear before detecting position.



\begin{thebibliography}{99}

\bibitem{c1} Zhang, Yifei. "A better autoencoder for image: Convolutional autoencoder." ICONIP17-DCEC. Available online: http://users. cecs. anu. edu. au/Tom. Gedeon/conf/ABCs2018/paper/ABCs2018\_paper\_58. pdf (accessed on 23 March 2017). 2018.
\bibitem{c2} Bank, Dor, Noam Koenigstein, and Raja Giryes. "Autoencoders." arXiv preprint arXiv:2003.05991 (2020).
\bibitem{c3} Badr, Will. “Auto-Encoder: What Is It? And What Is It Used for? (Part 1).” Medium, 9 Dec. 2021, towardsdatascience.com/auto-encoder-what-is-it-and-what-is-it-used-for-part-1-3e5c6f017726.
\bibitem{c4} Dataman, Chris Kuo/Dr. “Convolutional Autoencoders for Image Noise Reduction.” Medium, 8 Sept. 2022, towardsdatascience.com/convolutional-autoencoders-for-image-noise-reduction-32fce9fc1763.
\bibitem{c5} Zhang, Chen, et al. "Optimizing FPGA-based accelerator design for deep convolutional neural networks." Proceedings of the 2015 ACM/SIGDA international symposium on field-programmable gate arrays. 2015.
\bibitem{c6 }Xu, Jiayang, and Karthik Duraisamy. "Multi-level convolutional autoencoder networks for parametric prediction of spatio-temporal dynamics." Computer Methods in Applied Mechanics and Engineering 372 (2020): 113379.
\bibitem{c7} Turchenko, Volodymyr, Eric Chalmers, and Artur Luczak. "A deep convolutional auto-encoder with pooling-unpooling layers in caffe." arXiv preprint arXiv:1701.04949 (2017).
\bibitem{c8} Chen, Min, et al. "Deep feature learning for medical image analysis with convolutional autoencoder neural network." IEEE Transactions on Big Data 7.4 (2017): 750-758.
\bibitem{c9} Koushik, Jayanth. "Understanding convolutional neural networks." arXiv preprint arXiv:1605.09081 (2016).
\bibitem{c10} Wang, Wei, et al. "Generalized autoencoder: A neural network framework for dimensionality reduction." Proceedings of the IEEE conference on computer vision and pattern recognition workshops. 2014.
\bibitem{c11} Gokhale, Vinayak, et al. "A 240 g-ops/s mobile coprocessor for deep neural networks." Proceedings of the IEEE conference on computer vision and pattern recognition workshops. 2014.
\bibitem{c12} Chakradhar, Srimat, et al. "A dynamically configurable coprocessor for convolutional neural networks." Proceedings of the 37th annual international symposium on Computer architecture. 2010.
\bibitem{c13} Woods, Roger, et al. FPGA-based implementation of signal processing systems. John Wiley \& Sons, 2008.
\bibitem{c14} Camposano, Raul, and Jörg Wilberg. "Embedded system design." Design Automation for Embedded Systems 1.1 (1996): 5-50.
\bibitem{c15} Polzer, Andreas, et al. "Managing complexity and variability of a model-based embedded software product line." Innovations in Systems and Software Engineering 8.1 (2012): 35-49.
\bibitem{c16} Kattenborn, Teja, et al. "Review on Convolutional Neural Networks (CNN) in vegetation remote sensing." ISPRS Journal of Photogrammetry and Remote Sensing 173 (2021): 24-49.
\bibitem{c17} Benzigar, Mercy R., et al. "Advances on emerging materials for flexible supercapacitors: current trends and beyond." Advanced Functional Materials 30.40 (2020): 2002993.
\bibitem{c18} Sass, Ronald, and Andrew G. Schmidt. Embedded systems design with platform FPGAs: principles and practices. Morgan Kaufmann, 2010.
\bibitem{c19} Trimberger, Stephen M. Steve. "Three ages of fpgas: a retrospective on the first thirty years of fpga technology: this paper reflects on how moore's law has driven the design of fpgas through three epochs: the age of invention, the age of expansion, and the age of accumulation." IEEE Solid-State Circuits Magazine 10.2 (2018): 16-29.

\end{thebibliography}
\end{document}